# Vision-based Obstacle Removal System for Autonomous Ground Vehicles Using a Robotic Arm

Khashayar Asadi, SM.ASCE[1], Rahul Jain[2], Ziqian Qin[3], Mingda Sun[3], Mojtaba Noghabaei, SM.ASCE[4], Jeremy Cole[5], Kevin Han, Ph.D. M.ASCE[4], and Edgar Lobaton, Ph.D.[5]

[1] Department of Civil, Construction, and Environmental Engineering, North Carolina State University, 2501 Stinson Dr, Raleigh, NC 27606; email: kasadib@ncsu.edu
[2] Department of Civil and Environmental Engineering, Indian Institute of Technology Patna Bihta, Patna, Bihar 801103, India
[3] College of Computer Science and Technology, Zhejiang University, 38 Zheda Rd, Hangzhou, China
[4] Department of Civil, Construction, and Environmental Engineering, North Carolina State University, 2501 Stinson Dr, Raleigh, NC 27606
[5] Department of Electrical and Computer Engineering, North Carolina State University, 890 Oval Drive, Raleigh, NC 27606

## ABSTRACT

Over the past few years, the use of camera-equipped robotic platforms for data collection and visually monitoring applications has exponentially grown. Cluttered construction sites with many objects (e.g., bricks, pipes, etc.) on the ground are challenging environments for a mobile unmanned ground vehicle (UGV) to navigate. To address this issue, this study presents a mobile UGV equipped with a stereo camera and a robotic arm that can remove obstacles along the UGV's path. To achieve this objective, the surrounding environment is captured by the stereo camera and obstacles are detected. The obstacle's relative location to the UGV is sent to the robotic arm module through Robot Operating System (ROS). Then, the robotic arm picks up and removes the obstacle. The proposed method will greatly enhance the degree of automation and frequency of data collection for construction monitoring. The proposed system is validated through two case studies. The results successfully demonstrate the detection and removal of obstacles, serving as one of the enabling factors for developing an autonomous UGV with various construction operating applications.

## INTRODUCTION

The number of applications of automated platforms to work with human workers on construction sites has exponentially grown in the past few years. They are used for various activities, including floor cleaning (Prabakaran et al. 2018), wall building (Gosselin et al. 2016; Yu et al. 2009), wall painting (Sorour et al. 2011), data collection (Asadi et al. 2018a, 2019a; b; Asadi and Han 2018; Boroujeni and Han 2017), constructibility assessment (Balali et al. 2018; Noghabaei et al. 2019) and inspection (Menendez et al. 2018). However, the implementation of robots on construction sites is still limited. As more cost-effective applications are found, their use in practice will increase (Shakeri et al. 2015). Repetitive building tasks have been a target for automation studies (García de Soto et al. 2018). These studies mostly benefit from a robotic arm to handle different tasks.

(Sorour et al. 2011) developed a low weight autonomous robotic arm which was capable of painting the vertical wall. However, the robotic arm had only two degrees of freedom (DoF). Therefore, to increase the dimensional ability of robotic arm, (Gosselin et al. 2016) developed a robotic arm which had six degree of freedom to deposit layer by layer construction material (e.g., cement mortar) to build a multifunctional concrete wall. Recently, (García de Soto et al. 2018) developed a robotic system which was capable of building a curved concrete wall with mesh mold with the help of digital fabrication technique. Likewise, (Lublasser et al. 2018) developed a robot to make the foam concrete surface on the walls to gain facade finish.

There are many activities during the construction process which requires lifting of a three-dimensional object and placing at a different location. The above-mentioned studies have the ability to help during the construction stage in many ways; however, these robots don't have the capability of picking and keeping three-dimensional objects. Therefore, many research studies have tried to develop a robotic system which can perform such activities during the construction stage. (Skotheim et al. 2012) presented a stationary robotic system that scan and localize work pieces using a laser triangulation sensor for picking and placing operations. Likewise, (Furrer et al. 2017) conducted a study on a stationary robotic arm to utilize irregular materials found on-site for autonomous construction. They demonstrated that the robot was able to form a tower from the detected object by staking over each other. The main limitation with this method was the necessity of an offline step for scanning the objects and their geometry, which limits the robot's capability in dynamic scenes with rapid changes such construction sites.

To address this limitation, the objective of the presented study in this paper is to develop a robotic system that uses a Kinova Jaco arm (KINOVA 2008) to automatically grab different types of objects (e.g., pipe and brick for this study) and put them elsewhere. This automated task has different applications within the construction industry. For instance, removing obstacles along the UGV's path which prevents the robot to change its path and choose a longer path during a data collection with a predetermined path. Another example is material handling, moving a pile of objects from one location to another, which is a repetitive task that can be automated. To achieve this goal, a mobile robotic platform that builds on the authors' previous study is used. The platform is built on the Clearpath's Husky mobile robotic platform ("ClearPathRobotics:Husky" 2018). A laptop is used as a processing unit. A stereo camera is used as a visual sensor to provide depth information alongside the RGB images (i.e., images with red, green, and blue color channels). For ease of data exchange among various modules (Control, Context Awareness, Geometry Descriptor, and Robotic Arm), Robot Operating System (ROS) is used (Quigley et al. 2009).

**METHOD**

This section describes the proposed vision-based obstacle removal system. Figure 1 illustrates the integration among multiple modules. The Context Awareness Module receives images from a ZED stereo camera (StereoLabs 2018). A scene segmentation scheme processes the images and detects objects of interest (bricks and pipes) in the image by creating a mask around them. The segmented images and their corresponding depth information are inputs for the Geometry Descriptor Module. This module calculates the coordinate of the mask's center with respect to the arm coordinate system and compares it with the predetermined range that the robotic arm can access. If the object is in that range, a command is sent to the Control Module to stop the robot and then the mask's center coordinate alongside with the mask's direction is sent to the robotic arm. The arm can grab the objects in its range with the known location (coordinate of the mask center) and orientation (mask's direction). Depending on the application, the grabbed object is then located in another location. In this study, a joystick is used to send control commands to the Raspberry Pi (Control Module) that is connected to the robot platform. The main contribution of this approach is to integrate the stereo camera, context awareness, control, and robotic arm modules for developing an autonomous obstacle removal system. The capabilities of these modules are detailed in the following subsections.

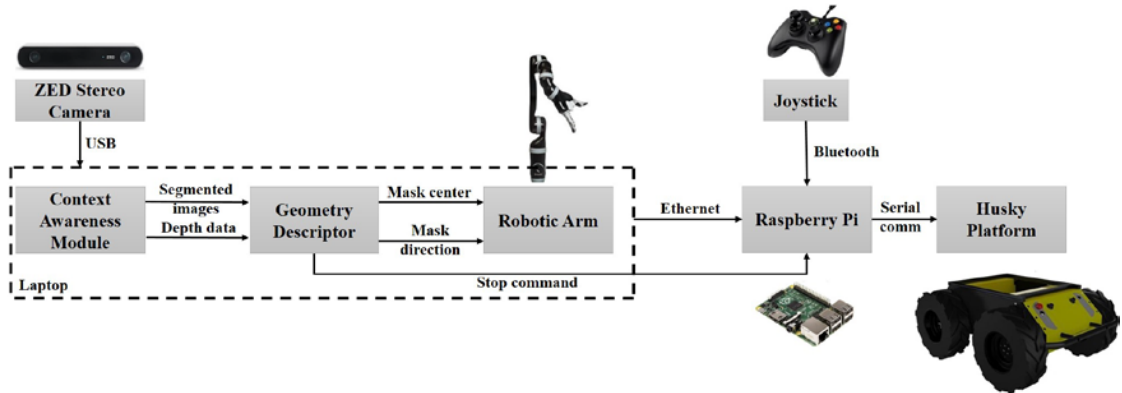

**Figure 1. An overview of the proposed vision-based obstacle removal system**

**ZED Stereo Camera**
The stereo camera provides the Context Awareness Module with RGB-D images (i.e., images with red, green, and blue color channels and the related depth value for each pixel). Depth values are respect to the camera coordinate system. To transfer all the values to the arm coordinate system, a rigid transformation matrix consists of translation and rotation is calculated. This matrix is calculated once during the system setup. Equation 1 shows this transformation where, PA and PC are the same physical points, described in arm and camera coordinate systems respectively. $t_C^A$ is a translation vector between the arm origin and the origin of the camera coordinate system. $R_C^A$ is the 3×3 rotation matrix of the camera axis with respect to the arm axis.

$$P^A = R_C^A + P^C + t_C^A \qquad (1)$$

The camera is placed on top of the robot platform with a fixed top-down view. Fixing the camera in top-down view has the following advantages: 1) the depth values are more accurate compared to any other camera angles. The reason is that all the objects in top-down view are less than 1.5 meters far from the camera, 2) this view includes objects that are close enough to the arm to be picked up, which prevent the unnecessary process of detecting objects that are not in the arm range, 3) The transformation matrix is calculated easier and it is more accurate compared to a camera with a random orientation.

**Context Awareness**
The Context Awareness Module receives images and corresponding depth information from the ZED stereo camera. The segmentation model proposed by (Asadi et al. 2019a) is used as a pixel-wise semantic segmentation method. This model takes pixel-wise labeled information as input for training. In the current study, the authors have collected and labeled about 1000 image frames. The following three classes have been chosen, as shown in Figure 2: brick, pipe, and unlabeled. To increase the number of training images, and preventing overfitting label-preserving data augmentation similar to (Krizhevsky et al. 2012) is used. Different methods of data augmentation such as flipping, random color jitter, and random Gaussian noise are implemented and generate 32 new images from each image. The network is trained in two steps. The first step involves training of the encoder part, which provides a label map of size 64×32. In the second step, the decoder is trained with the encoder to upscale the intermediate map to full image dimensions. With pixel-wise semantic segmentation, the object of interest can be easily detected from the resulting segmented image (see Figure 2).

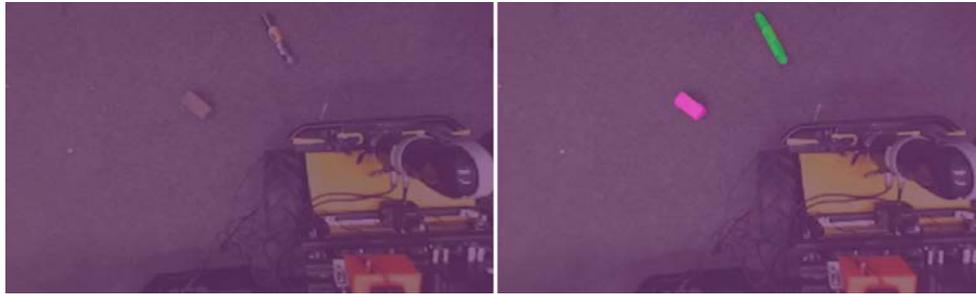

**Figure 2. Example of an image received from the left lens of the ZED camera (left) and the result segmented image (right)**

**Geometry Descriptor**

The Geometry Descriptor Module calculates the coordinate of the detected mask's center (see ($x_c$, $y_c$, $z_c$) in the left image of Figure 3) and its direction with respect to the arm coordinate system. This information is necessary for the arm to grab the object properly. The calculated mask center is compared with the predetermined range that the robot is able to reach. If the object is in the range, a command is sent to the Raspberry Pi to stop the robot. Then, the direction of the mask is calculated based on the depth values of the detected mask. For this purpose, the longest edge of the object is determined and its slope is calculated (see the red line in Figure 3). This slope represents the direction of the object with respect to the arm coordinate system. The arm's hand is supposed to reach the top of the object in parallel to this direction (to be further detailed in the following subsection).

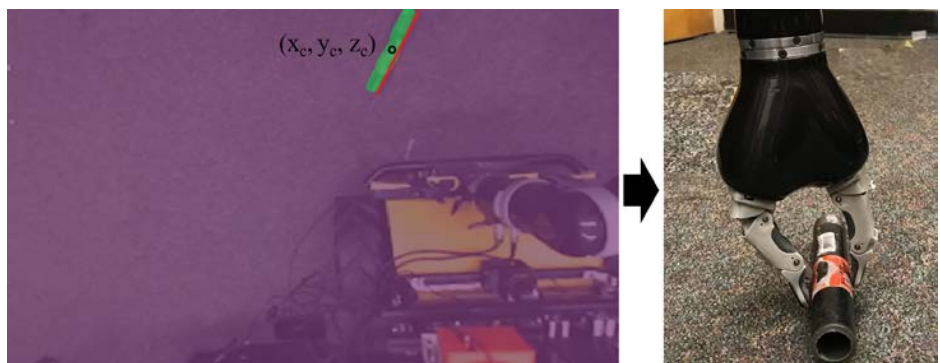

**Figure 3. The mask's center (black dot) and direction (red line) are calculated. By this information, the arm can grab the object parallel to the object's longest edge (right image).**

**Robotic Arm**

Kinova Jaco robotic arm (KINOVA 2008) with six DoF is used for this experiment. As previously mentioned, the arm is able to grab the object within a predetermined range. This range is calculated based on the arm's constraints (Palacios 2015) in different directions. The Robotic Arm Module receives the location of the object (center of the detected mask) that is within the arm range alongside with the object's direction (i.e., the slope of the longest edge of the mask) and then moves to reach the top of the object with a proper pose. The arm movement follows specific motions in order. The first motion starts from an initial pose (arm's home position). The arm moves to locate on top of the object. In the next movement, it goes down to reach the object and grab it. Finally, the arm handles the object and releases it in another location (depends on the application). The average time for the proposed system to grab an object is almost 20 seconds. The response times for the Context Awareness and Geometry Descriptor Modules are negligible which are less than a second.

**EXPERIMENTAL SETUP AND RESULTS**

The proposed system was tested with 10 objects from two classes (five bricks and five pipes). These objects varied in terms of size and their directions in the scene. To test the system in different light condition, the experiment held in an indoor environment. The segmentation model ran at a speed of 21 fps for an input image size of 512×256 on the laptop with the following specification: 16 GB DDR3 RAM, Intel Core i7-4710HQ quad-core Haswell processor, and NVIDIA GeForce GTX 960M. The system grabbed seven objects successfully and failed to grab one pipe and two bricks. Table 1 shows the failed cases besides the modules that caused the failure.

**Table1. Causes of failures**

| Failed Cases | Stereo ZED Camera | Context Awareness | Geometry Descriptor | Robotic Arm |
|---|---|---|---|---|
| Brick | X | - | - | - |
| Brick | - | X | X | - |
| Pipe | - | - | - | X |

Since the proposed method uses a ZED stereo camera for estimating depth values, accurate estimation of depth is vital for this method. However, in the first failure, the estimated location of the brick had the error of 2cm. By processing the output values from each module, it could be observed that the cause of this error was inaccurately estimated depth from the camera.

In the second failure, the Context Awareness Module failed to create a complete mask around the object (see the left image in Figure 4. This happens because of a noticeable change in the light condition. Consequently, the Geometry Descriptor also failed to calculate the center of the object accurately from the imperfect detected mask. So, the arm reached the object, but It failed to grab it from the proper pose which resulted in a failure (see the right image in Figure 4). Training the segmentation model with more data in different light conditions can solve this issue.

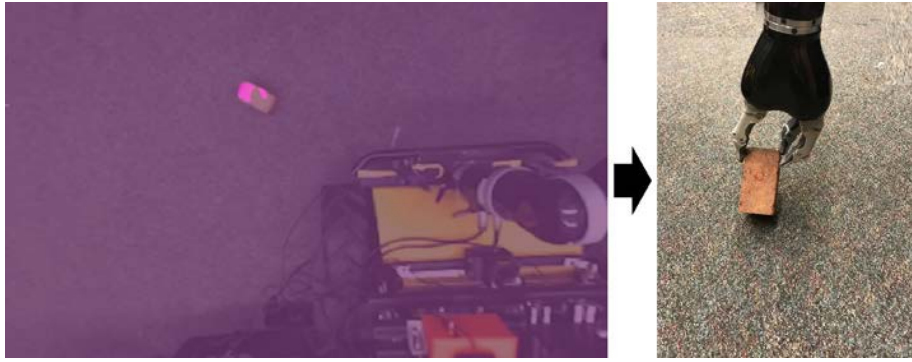
**Figure 4. Example of imperfect mask resulting in the second failure.**

By processing the last failure, it could be observed that although the received information from the Camera, Context Awareness, and the Geometry Descriptor Modules were all correct, the arm failed to reach the object by following the predetermined orders. For instance, instead of placing on top of the object and then moving down to grab the object, the arm can reach the object properly if it first moves a little down and then moves to the top of the object. So, depending on the object's location with respect to the arm coordinate system, different scenarios of picking the object can be determined to solve this issue.

**CONCLUSION**

This paper presents a vision-based obstacle removal system for autonomous ground robots using a Kinova Jaco robotic arm. A scene segmentation pipeline in integration with a stereo camera is used to detect the objects of interests. Then, the Geometry Descriptor Module tracks the location and orientation of the detected objects. This information is sent to the robotic arm to move to the object, grab it, and place it to a desirable location. The system validated through an experiment in an indoor environment. By investigating on cases with failure, the probable solutions for better performance of the system were suggested. The proposed system has the potential for enabling computer vision systems for object handling for automated construction applications.